\documentclass[preprint, 12pt]{elsarticle}

\usepackage{graphicx}
\usepackage{amsmath,amssymb,mathtools}
\usepackage{hyperref} 
\usepackage{booktabs}
\usepackage{multirow}
\usepackage{float}
\usepackage{subfig}
\usepackage{algorithm}             
\usepackage[noend]{algpseudocode}  
\usepackage{makecell}
\usepackage{adjustbox}

\journal{arXiv}

\begin{document}

\begin{frontmatter}

\title{Next-Depth Lookahead Tree}

\author[gachon_a]{Jaeho Lee}
\ead{loper00@gachon.ac.kr}

\author[gachon_b]{Kangjin Kim\corref{cor1}}
\ead{kim0616@gachon.ac.kr}

\author[gachon_c]{Gyeong Taek Lee\corref{cor1}}
\ead{leegt@gachon.ac.kr}

\address[gachon_a]{Department of Mathematical Finance, Gachon University, Seongnam 13120, Republic of Korea}
\address[gachon_b]{Department of Applied Statistics, Gachon University, Seongnam 13120, Republic of Korea}
\address[gachon_c]{Department of Mechanical, Smart, and Industrial Engineering, Gachon University, Seongnam 13120, Republic of Korea}

\cortext[cor1]{These authors contributed equally to this work. \\ Corresponding authors}

\begin{abstract}
This paper proposes the \emph{Next-Depth Lookahead Tree} (NDLT), a single-tree model designed to improve performance by evaluating node splits not only at the node being optimized but also by evaluating the quality of the next depth level. Conventional decision trees (DTs) rely on greedy node-by-node partitioning, which often fails to ensure global optimality of the tree and is prone to local optima when early splits are suboptimal. To overcome this limitation, NDLT employs a next-depth lookahead strategy that jointly considers the immediate impurity reduction at the parent node and the expected impurity reduction at its child nodes. Empirical evaluation on diverse and complex datasets, including high-dimensional and imbalanced cases, demonstrates that NDLT achieves performance comparable to or better than classical DTs and ensemble models such as Random Forests, XGBoost, and LightGBM. These results show that NDLT preserves the interpretability of a single tree while delivering robust predictive accuracy, making it an effective approach for real-world applications where both transparency and performance are required.
\end{abstract}


\end{frontmatter}

\section{Introduction} \label{sec:Introduction}
Decision Tree (DT) represents one of the earliest and most established algorithms in the field of machine learning, valued for its simplicity, interpretability, and broad applicability \cite{izza2020explain}. As a white-box model, a DT generates predictions by applying explicit rules and conditions to input data, ensuring transparency in the decision-making process and enabling straightforward interpretation of results \cite{costa2023survey}. These attributes not only facilitate the identification of key independent variables for subsequent analysis but also support the development of more parsimonious models \cite{kazemitabar2017variable}. Owing to these advantages, decision trees have been extensively employed across diverse domains, ranging from life sciences—including genetics, clinical medicine, and bioinformatics—to organizational contexts such as cybersecurity, personnel management, turnover prediction, and manufacturing \cite{ahmed2023turnover,wang2024fruit,amruth2024cloud,choi2017manufacturing}. Structurally, a DT models the decision-making process by recursively partitioning data into subsets through a hierarchical organization of nodes, consisting of a root node, internal nodes, and terminal leaf nodes, with each partition determined by specific conditions \cite{strobl2009recursive}.

However, the DT has several limitations. As the data size and the number of observations increase, the number of possible DTs becomes virtually infinite. This is because the number of possible splitting combinations grows exponentially with the number of features and observations \cite{kozak2019dt}. Consequently, finding an optimal DT is considered NP-hard—not only is determining the optimal tree structure NP-hard, but verifying the optimality of a given tree is also NP-hard \cite{olaru2003fuzzy}. Since every possible splitting candidate must be evaluated for each feature, the process becomes inefficient and computationally expensive for large datasets \cite{rouhi2020feature}. Moreover, there is a high risk of converging to a local optimum during this process. Although increasing the tree depth drives the training error toward zero, it also heightens the risk of overfitting \cite{costa2023survey}. To mitigate these problems, various ensemble tree methods have been proposed.

To address these limitations, we propose the Next-Depth Lookahead Tree (NDLT), a novel extension of the conventional decision tree framework. Unlike traditional DT algorithms that adopt a purely greedy strategy by selecting the best split at the current node without further foresight, the NDLT employs what we define as the next-depth lookahead strategy. This strategy explicitly evaluates each candidate split not only by its immediate impurity reduction but also by the prospective impurity improvement achieved at the subsequent child nodes. In other words, the splitting decision at a given node is informed by anticipating the quality of the next depth level. By integrating this forward-looking criterion, the NDLT mitigates the risk of suboptimal early splits, reduces the likelihood of converging to local optima, and promotes a more globally consistent tree structure. This design preserves the interpretability of classical decision trees while enhancing robustness and predictive accuracy, particularly in high-dimensional or complex datasets.

\section{Methodology} \label{sec:methodology}

\subsection{Overview}

\begin{figure}[ht!]
    \centering
    \includegraphics[width=0.75\linewidth]{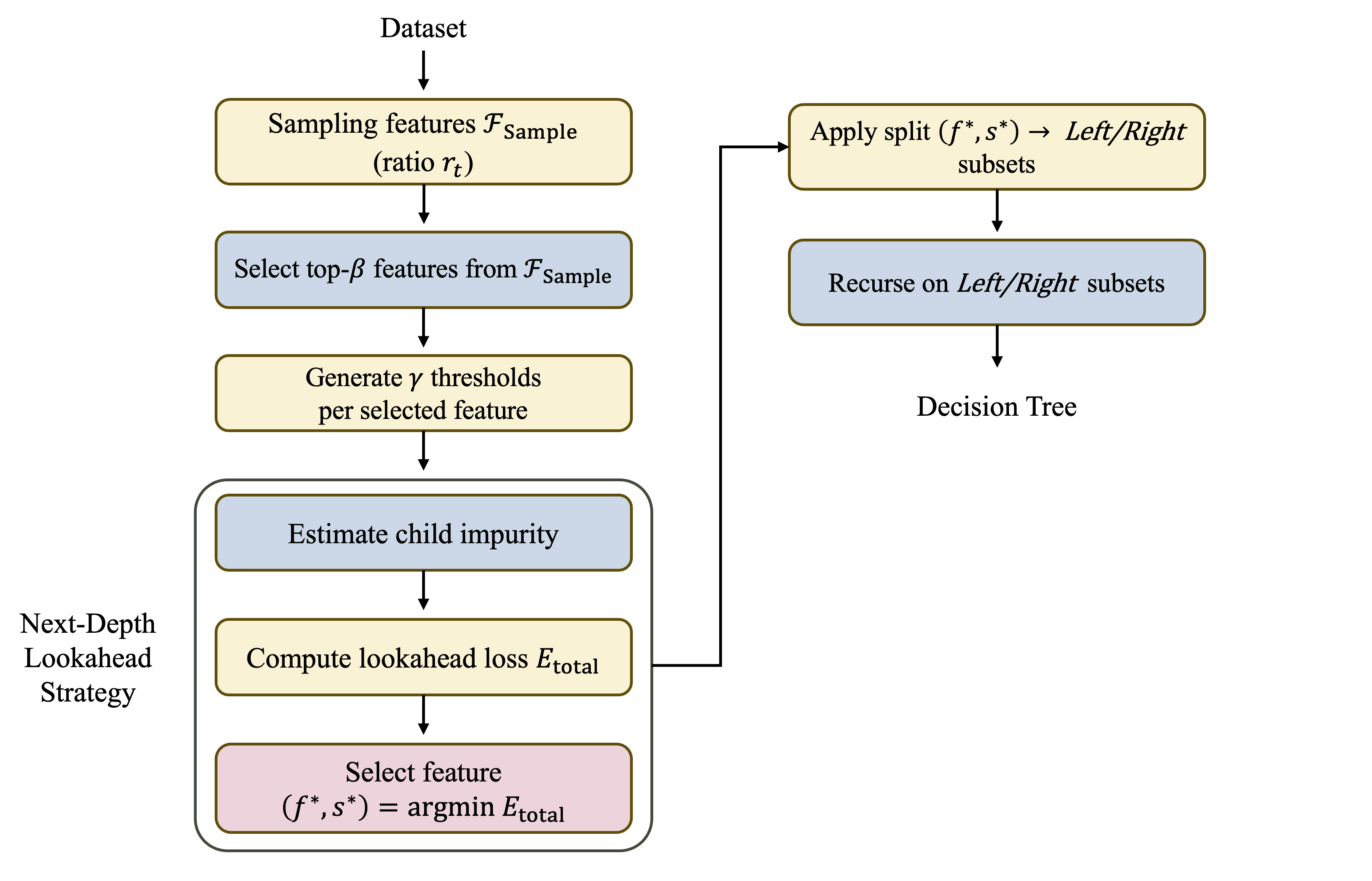}
    \caption{NDLT model architecture}
    \label{fig:NDLT_flowchart}
\end{figure}

Figure \ref{fig:NDLT_flowchart} shows the overall process of the NDLT model. The proposed NDLT model takes input data $X$ with corresponding labels $y$. The selection of optimal splits at each internal node is governed by a probabilistic feature sampling strategy coupled with multi-threshold evaluation. Given a node with input samples $(X, y)$, a random subset of features $\mathcal{F}_{\text{sample}} \subset \{1,2, \dots, d\}$ is selected with sampling ratio $r_t$, where $d$ denotes the total number of features. For each feature $f \in \mathcal{F}_{\text{sample}}$, let $m_{f} = \min(\gamma, \: u_{f})$. Then candidate thresholds $\{s_{1}, s_{2}, \dots, s_{m_{f}}\}$ are generated from quantiles or midpoints of the sorted values of $x_{f}$, where $u_f$ denotes the number of admissible thresholds 
determined by the unique values of $x_f$.

At each internal node, a feature $f$ and a threshold $s$ are selected to partition the data into two disjoint subsets:
\[
    \text{Left node: } \{x \in X \mid x_f \leq s\}, \quad
    \text{Right node: } \{x \in X \mid x_f > s\}.
\]
Each candidate split $(f, \: s)$ partitions the data into $X_{\text{left}}$ and $X_{\text{right}}$ and their respective Gini impurities, $G_L$ and $G_R$, are computed. The impurity of the split is then assessed using the weighted average:
\begin{equation}
    G_{\text{upper}} = \frac{n_L}{n} G_L + \frac{n_R}{n} G_R,
\end{equation}
where $n = n_L + n_R$ is the number of samples at the node, $n_L$ is the number of samples in the left child, and $n_R$ is the number of samples in the right child.
where $n = n_L + n_R$ is the number of samples at the node, $n_L $ is the number of samples in the left child, and $n_R $ is the number of samples in the right child.

    
\subsection{Function Definitions}
To keep notation concise, we introduce helper functions and refer to them throughout instead of repeating intermediate impurity calculations.
\begin{align*}
\text{\textsc{Gini}}(y)                         &:= 1 - \sum_{c} p_c^{2}, \quad p_c = \frac{1}{\lvert y\rvert}\sum_{i}\mathbb{I}\{y_i=c\},\\[4pt]
\text{\textsc{SplitImpurity}}(X_L,y_L,X_R,y_R)  &:= \frac{n_L}{n}\,\text{\textsc{Gini}}(y_L) + \frac{n_R}{n}\,\text{\textsc{Gini}}(y_R), \ \\[4pt]
\text{\textsc{LowerEval}}(X_L,y_L,X_R,y_R)      &:= \min\!\bigl(G_L^{\min},\,G_R^{\min}\bigr) + \tfrac{1}{2}\!\left(G_L^{\min}+G_R^{\min}\right),  \\[4pt]
w_1(d,\bar{e})                                   &:= (1-\bar{e})\,\delta^{d},  \\[4pt]
E_{\text{total}}                                & := \begin{aligned}[t]
                                                       &G_{\text{upper}}\cdot w_1 w_2 \\ 
                                                       & \qquad + \bigl(G_{\text{lower}}+\varepsilon\bigr)\,(1-w_1)\,(1-w_2). 
                                                    \end{aligned}
\end{align*}

Here $\mathbb{I}$ is the indicator function, $\bar{e}$ is the mean $G_{\text{upper}}$ over retained candidates at the current node, $d$ is the depth, $\delta\in(0,1)$ is a decay factor, $w_2\in[0,1]$ balances upper and lower terms, and $\varepsilon>0$ ensures numerical stability. We use $(f,s)$ for \emph{parent} candidates evaluated at the current node and $(f',s')$ for \emph{child-side} inner candidates considered in the one-step localized search.

Given a parent candidate $(f,s)$, the node data $(X,y)$ are split into $(X_L,y_L)$ and $(X_R,y_R)$ and $G_{\text{upper}}$ is computed by \textsc{SplitImpurity}. For each child side $S\in\{L,R\}$, we perform a one-step localized search restricted to the retained top-$\beta$ feature pool $\mathcal{F}_{\text{top-}\beta}$ from the parent node. We sample, without replacement, a side-specific subset $\mathcal{F}^{(S)}\subseteq \mathcal{F}_{\text{top-}\beta}$ at rate $r_t$. For each $f'\in\mathcal{F}^{(S)}$, let $u_{f',S}$ denote the number of admissible thresholds determined by the sorted support of $x_{f'}$ on $X_S$ and define $m_{f',S}=\min(\gamma,\,u_{f',S})$. From that support, construct up to $m_{f',S}$ thresholds $s'\in\mathcal{T}^{(S)}_{f'}$ (midpoints between consecutive unique values or designated quantiles). Each inner pair $(f',s')$ induces a side-split

\begin{align*}
    (X_{S,L}^{(f',s')},y_{S,L}^{(f',s')}) & =\{(x_i,y_i)\in(X_S,y_S): x_{i,f'}\le s'\}, \\[3pt]
    (X_{S,R}^{(f',s')},y_{S,R}^{(f',s')}) & =\{(x_i,y_i)\in(X_S,y_S): x_{i,f'}> s'\}.
\end{align*}
We enforce the minimum leaf constraint by admitting only candidates with
$|X_{S,L}^{(f',s')}|\ge {min\_samples\_leaf}$ and
$|X_{S,R}^{(f',s')}|\ge {min\_samples\_leaf}$.
For each admissible inner split we evaluate
\[
G_S(f',s')\;:=\;\textsc{SplitImpurity}\!\big(X_{S,L}^{(f',s')},y_{S,L}^{(f',s')},\,X_{S,R}^{(f',s')},y_{S,R}^{(f',s')}\big),
\]
and define the side-wise minimum
\[
G_S^{\min}\;:=\;
\begin{cases}
\min\limits_{(f',s')\ \text{admissible}}\, G_S(f',s'), & \text{if admissible candidates exist},\\[6pt]
0, & \text{otherwise (stability default).}
\end{cases}
\]
Applying this to $S=L$ and $S=R$ yields $G_{L}^{\min}$ and $G_{R}^{\min}$, which are aggregated by \textsc{LowerEval} and then combined with $G_{\text{upper}}$ in $E_{\text{total}}$.

\subsection{Incorporating Lower Node Evaluation}
Conventional DT algorithms determine each node using a greedy splitting strategy in a top-down manner; however, repeating this top-down procedure across the tree does not necessarily yield an optimally structured tree. For example, even if the root split is not the most impurity-reducing choice, appropriate feature–threshold selections at its child nodes can further reduce the weighted Gini impurity, yielding a lower overall impurity than any single-step decision at the root. To address this limitation, the proposed NDLT model assesses the discriminative potential of the child nodes, thereby enabling the tree to eventually achieve superior overall performance.

For each candidate pair $(f, s)$ originating from the previously retained top-$\beta$ features, the dataset is partitioned into $X_{\text{left}}$ and $X_{\text{right}}$ by the binary rule. Within each partition, a secondary feature subset (left and right, respectively) is resampled from the same top-$\beta$ pool using the sampling ratio $r_t$. A localized greedy search over candidate thresholds (derived from quantiles or midpoints of the sorted unique values) is then conducted to estimate the minimum attainable weighted Gini impurity in that partition. Let $G_L^{\min}$ and $G_R^{\min}$ denote the resulting minima for the left and right subsets, respectively. If no valid secondary splits are found for a side, its estimate is defaulted to zero to preserve numerical stability and avoid over-penalization. The aggregated lower-node score is
\begin{equation}
  G_{\text{lower}} \;=\; \min(G_L^{\min}, G_R^{\min}) \;+\; \tfrac{1}{2}\big(G_L^{\min}+G_R^{\min}\big).
\end{equation}

\subsection{Total Error Score and Weighted Decision}
To jointly evaluate the local effectiveness of a split and the prospective quality of its descendants, NDLT minimizes the composite objective
\begin{equation}
  E_{\text{total}} = G_{\text{upper}}\cdot w_1(d,\bar{e}) \cdot w_2 + \big(G_{\text{lower}}+\varepsilon\big)\big(1-w_1(d,\: \bar{e})\big)\big(1-w_2\big),
\end{equation}
where $w_1(d,\bar{e})=(1-\bar{e})\delta^{d}$ downweights the upper-node term with depth and with larger average impurity $\bar{e}$, while $w_2$ tunes the relative importance between upper and lower terms. The chosen split is
\begin{equation}
  (f^{*}, s^{*}) \in \arg\min_{(f,s)} E_{\text{total}}(f,s).
\end{equation}

\subsection{Algorithmic Description}
We present function-based pseudocode that calls the helper functions instead of repeating low-level impurity computations. Parent-level split candidates at the current node are denoted by $(f,s)$, while the inner one-step search performed inside each child uses $(f',s')$ to avoid notational collision. The helper functions implement (i) impurity calculation at a single split (\textsc{Gini}, \textsc{SplitImpurity}), (ii) the one-step localized search inside each child to obtain $G_{L}^{\min}$ and $G_{R}^{\min}$ (\textsc{LowerEval}, optionally factored via \textsc{ComputeSideMinImpurity}), and (iii) the depth-aware combination of upper/lower terms into the final score (\textsc{TotalError}).

The first helper block below provides a compact implementation of \textsc{LowerEval}: for each side $S\in\{L,R\}$ it re-samples features from the retained top-$\beta$ pool at rate $r_t$, generates up to $\gamma$ thresholds per sampled feature from the child’s support (midpoints or quantiles), enforces the minimum leaf-size constraint on both grandchildren, evaluates the weighted impurity of every admissible inner split, and records the side-wise minimum $G_S^{\min}$, falling back to $0$ when no admissible candidate exists. The returned value aggregates both sides as $\min(G_L^{\min},G_R^{\min})+\tfrac{1}{2}(G_L^{\min}+G_R^{\min})$ to reflect best-case potential and an averaged tendency for further separability.

For clarity and reuse, the second helper block factors the inner search into \textsc{ComputeSideMinImpurity}. Given a child $(X_S,y_S)$, it immediately returns $0$ if $|X_S|<2\cdot min\_samples\_leaf$ (no feasible inner split), otherwise it samples $\mathcal{F}^{(S)}\subseteq\mathcal{F}_{\text{top-}\beta}$ without replacement at rate $r_t$, constructs at most $m_{f',S}=\min(\gamma,u_{f',S})$ thresholds per $f'$ from the child’s sorted support, filters candidates that violate the leaf-size constraint, and returns the minimum \textsc{SplitImpurity} among the admissible set (or $0$ if empty). The outer \textsc{LowerEval} then calls this routine for $S=L$ and $S=R$ and applies the same aggregation. Both helper blocks are equivalent by construction; the factored version makes the admissibility checks and the cap $m_{f',S}=\min(\gamma,u_{f',S})$ explicit.

Finally, the full \emph{Next-Depth Lookahead Tree} procedure (kept unchanged) first samples features to form $\mathcal{F}_{\text{sample}}$, builds up to $m_f=\min(\gamma,u_f)$ thresholds per feature $f$, and evaluates the parent-level impurity $G_{\text{upper}}$ for every $(f,s)$. Per-feature best candidates (with minimal $G_{\text{upper}}$) are retained to rank features; the top-$\beta$ set and its mean impurity $\bar{e}$ are recorded. For each retained $(f,s)$, the data are re-split, $G_{\text{upper}}$ is recomputed, $G_{\text{lower}}$ is obtained via \textsc{LowerEval} (hence $G_L^{\min}$, $G_R^{\min}$), and the depth-aware total score
\begin{align*}
    E_{\text{total}}    & = G_{\text{upper}}\cdot w_1(d,\bar{e})\,w_2+\big(G_{\text{lower}}+\varepsilon\big)\big(1-w_1(d,\bar{e})\big)\big(1-w_2\big), \\[3pt]
    w_1(d,\bar{e})      & = (1-\bar{e})\delta^{d},
\end{align*}
is used to select $(f^{*},s^{*})\in\arg\min E_{\text{total}}(f,s)$, provided both children satisfy the minimum leaf-size constraint; otherwise the node becomes a leaf with the majority label.

\begin{algorithm}[H]
    \caption{Helper Functions for NDLT}
    \scalebox{0.9}{
    \begin{minipage}{1.1\textwidth}
    \begin{algorithmic}[1]
    \Function{Gini}{$y$}
      \State Compute class proportions $\{p_c\}$ from $y$
      \State \Return $1 - \sum_c p_c^2$
    \EndFunction
    
    \Function{SplitImpurity}{$X_L,y_L,X_R,y_R$}
      \State $n_L \gets |X_L|,\; n_R \gets |X_R|,\; n \gets n_L+n_R$
      \State \Return $\frac{n_L}{n}\,\Call{Gini}{y_L} + \frac{n_R}{n}\,\Call{Gini}{y_R}$
    \EndFunction
    
    \Function{ComputeSideMinImpurity}{$X_S,y_S;\; r_t,\gamma,\beta,min\_samples\_leaf$}
      \If{$|X_S| < 2\cdot min\_samples\_leaf$}
         \State \Return $0$ \Comment{no valid inner split can satisfy both sides}
      \EndIf
      \State Sample $\mathcal{F}^{(S)} \subseteq \mathcal{F}_{\text{top-}\beta}$ at rate $r_t$ \Comment{without replacement}
      \State $G_{\min}\gets +\infty$
      \For{$f' \in \mathcal{F}^{(S)}$}
         \State $u_{f',S}\gets$ \# admissible thresholds from unique sorted values of $x_{f'}$ on $X_S$
         \State $m_{f',S}\gets \min(\gamma, u_{f',S})$
         \State Build up to $m_{f',S}$ thresholds $\{s'\}$ (midpoints/quantiles on $X_S$)
         \For{each $s' \in \{s'\}$}
            \State Form $(X_{S,L}^{(f',s')},y_{S,L}^{(f',s')})$, $(X_{S,R}^{(f',s')},y_{S,R}^{(f',s')})$
            \If{$|X_{S,L}^{(f',s')}|<min\_samples\_leaf$ \textbf{ or } $|X_{S,R}^{(f',s')}|<min\_samples\_leaf$}
               \State \textbf{continue}
            \EndIf
            \State $G \gets \Call{SplitImpurity}{X_{S,L}^{(f',s')},y_{S,L}^{(f',s')},X_{S,R}^{(f',s')},y_{S,R}^{(f',s')}}$
            \If{$G < G_{\min}$} \State $G_{\min}\gets G$ \EndIf
         \EndFor
      \EndFor
      \State \Return $(G_{\min}=+\infty)\ ?\ 0:\,G_{\min}$
    \EndFunction
    
    \Function{LowerEval}{$X_L,y_L,X_R,y_R;\; r_t,\gamma,\beta$}
      \State $G_L^{\min} \gets \Call{ComputeSideMinImpurity}{X_L,y_L;\; r_t,\gamma,\beta,min\_samples\_leaf}$
      \State $G_R^{\min} \gets \Call{ComputeSideMinImpurity}{X_R,y_R;\; r_t,\gamma,\beta,min\_samples\_leaf}$
      \State \Return $\min(G_L^{\min},G_R^{\min}) + \tfrac{1}{2}(G_L^{\min}+G_R^{\min})$
    \EndFunction
    
    \Function{TotalError}{$G_{\text{upper}}, G_{\text{lower}}, \bar{e}, d, \delta, w_2, \varepsilon$}
      \State $w_1 \gets (1-\bar{e})\cdot \delta^{d}$
      \State \Return $G_{\text{upper}}\cdot w_1\cdot w_2 + (G_{\text{lower}}+\varepsilon)(1-w_1)(1-w_2)$
    \EndFunction
    \end{algorithmic}
    \end{minipage}
    }
\end{algorithm}

\begin{algorithm}[H]
    \caption{Next-Depth Lookahead Tree} 
    \scalebox{0.9}{
    \begin{minipage}{1.1\textwidth}
    \begin{algorithmic}[1]
    \Require training data $X$, labels $y$
    \State \textbf{Parameters}: max\_depth, min\_samples\_leaf, $r_t$ (feature sampling ratio), $\gamma$ (thresholds/feature), $\beta$ (top features), $\delta\in(0,1)$, $w_2\in[0,1]$, $\varepsilon>0$
    
    \Function{BuildTree}{$X,y,d$}
      \If{pure$(y)$ \textbf{ or } $|X|\le min\_samples\_leaf$ \textbf{ or } $d\ge max\_depth$}
        \State \Return majority$(y)$
      \EndIf
    
      \State Sample feature subset $\mathcal{F}_{\text{sample}}$ at ratio $r_t$
      \For{$f \in \mathcal{F}_{\text{sample}}$}
        \State $u_{f} \gets $ the number of admissible thresholds determined by the unique values of $x_{f}$
        \State $m_{f} \gets \min(\gamma, u_{f})$
        \State Generate up to $m_{f}$ thresholds $\{s\}$ (quantiles/midpoints)
        \For{each $s$}
          \State Split $X$ into $(X_L,y_L)$ and $(X_R,y_R)$ by $(f,s)$
          \State $G_{\text{upper}}(f,s) \gets \Call{SplitImpurity}{X_L,y_L,X_R,y_R}$
        \EndFor
        \State Keep top candidates per $f$ by \emph{minimum} $G_{\text{upper}}$
      \EndFor
    
      \State Rank features by their minimal $G_{\text{upper}}$; retain top-$\beta$
      \State $\bar{e} \gets$ mean of stored $G_{\text{upper}}$ values over retained candidates across top-$\beta$
    
      \For{each retained candidate $(f,s)$}
        \State Split $X \to (X_L,y_L),(X_R,y_R)$ by $(f,s)$
        \State $G_{\text{upper}} \gets \Call{SplitImpurity}{X_L,y_L,X_R,y_R}$
        \State $G_{\text{lower}} \gets \Call{LowerEval}{X_L,y_L,X_R,y_R;\; r_t,\gamma,\beta}$
        \State $E_{\text{total}}(f,s) \gets \Call{TotalError}{G_{\text{upper}},G_{\text{lower}},\bar{e},d,\delta,w_2,\varepsilon}$
      \EndFor
    
      \State $(f^{*},s^{*}) \gets \arg\min_{(f,s)} E_{\text{total}}(f,s)$
      \State Split $X \to (X_L,y_L),(X_R,y_R)$ by $(f^{*},s^{*})$
    
      \If{$|X_L|<min\_samples\_leaf$ \textbf{ or } $|X_R|<min\_samples\_leaf$}
        \State \Return majority$(y)$
      \EndIf
    
      \State \Return node with $(f^{*},s^{*})$ and children:
      \State \hspace{1.0em} $left \gets \Call{BuildTree}{X_L,y_L,d{+}1}$
      \State \hspace{1.0em} $right \gets \Call{BuildTree}{X_R,y_R,d{+}1}$
    \EndFunction
    
    \State \Return \Call{BuildTree}{$X,y,0$}
    \end{algorithmic}
    \end{minipage}
    }
\end{algorithm}

\subsection{Optimization}

In the fixed split regime, a predefined sequence of feature-threshold pairs $\{(f_d, s_d)\}_{d=0}^{D^{\ast}}$ is provided, where each pair corresponds to a specific depth $d$ in the tree. When the training process reaches depth $d$, the model bypasses dynamic candidate evaluation and applies the predefined split $(f_{d}, s_{d})$ directly. This allows the tree to encode prior knowledge.

The decision tree is constructed recursively, beginning from the root node and expanding by sequentially applying the optimal split at each node. At each stage of recursion, the algorithm determines whether to terminate the branch or to further divide the node based on 
stopping criteria.

If a fixed split $(f_d, s_d)$ is specified at the current depth $d$, it is applied directly. Otherwise, the model evaluates candidate splits according to the total error score $E_{\text{total}}$ and selects the optimal pair $(f^*, s^*)$ that minimizes this value. If no valid split is found, the node is assigned a class label corresponding to the majority class among its samples.

\section{Experiments}
\label{sec:experiments}

In this section, we present a series of empirical evaluations designed to validate the effectiveness of the proposed NDLT. 
The experiments are conducted on a diverse suite of binary classification datasets in order to ensure that the results are robust to differences in sample size, feature dimensionality, and class imbalance. 
Our primary objective is examining whether the next-depth lookahead strategy embedded in NDLT enables a single decision tree to achieve competitive or superior performance relative to conventional decision trees methods. 

\subsection{Datasets}

To comprehensively evaluate the proposed NDLT model, we employ thirteen publicly available binary classification datasets drawn from the UCI Machine Learning Repository.
These benchmarks encompass a broad range of sample sizes and feature dimensionalities, and they also exhibit varying degrees of class imbalance.
By covering such diverse characteristics, the experimental suite ensures that the assessment of NDLT is not limited to a particular data regime, but rather reflects performance robustness across heterogeneous application scenarios.

The public datasets are as follows:
\begin{itemize}
\item Bank Marketing dataset(Bank): Used to predict bank marketing campaign response; 45211 samples, 16 variables, label rate of 11.70\%.
\item Blood Transfusion Service Center dataset (Blood): Used to predict whether a donor will donate blood in the future; 748 samples, 4 variables, label rate of 23.80\%.
\item Breast Cancer Wisconsin Diagnostic dataset (Breast): Used to predict breast cancer diagnosis; 569 samples, 30 variables, label rate of 62.74\%.
\item Chronic Kidney Disease dataset (Kidney): Used to predict chronic kidney disease presence; 400 samples, 24 variables, label rate of 0.50\%.
\item Congressional Voting Records dataset (Voting): Used to classify party affiliation from voting records; 435 samples, 16 variables, label rate of 61.38\%.
\item Heart Failure Clinical Records dataset (Heart): Used to predict heart failure events from clinical records; 299 samples, 12 variables, label rate of 32.11\%.
\item Horse Colic dataset (Horse): Used to predict horse colic outcomes; 368 samples, 27 variables, label rate of 63.04\%.
\item Iranian Churn dataset (Iranian): Used to predict customer churn in telecom; 3150 samples, 13 variables, label rate of 15.71\%.
\item Ozone Level Detection dataset (Ozone): Used to predict surface ozone level; 5070 samples, 72 variables, label rate of 4.60\%.
\item Raisin dataset (Raisin): Used to classify raisin varieties; 900 samples, 7 variables, label rate of 50.00\%.
\item Regensburg Pediatric Appendicitis dataset (Regensburg): Used to predict appendicitis severity/treatment in pediatric cases; 782 samples, 53 variables, label rate of 0.13\%.
\item Rice Cammeo And Osmancik dataset (Rice): Used to classify rice varieties (Cammeo vs Osmancik); 3810 samples, 7 variables, label rate of 57.22\%.
\item Spambase dataset (Spambase): Used to detect spam emails; 4601 samples, 57 variables, label rate of 39.40\%.
\end{itemize}

\subsection{Experimental Design}

The experimental design of this study was carefully constructed to provide a fair, reproducible, and comprehensive evaluation of the proposed NDLT model in comparison with a wide range of baselines. We followed a multi-stage experimental pipeline consisting of preprocessing, the definition of evaluation metrics, the selection of baseline methods, and the specification of model configurations for the NDLT. Each component was implemented in a consistent framework so that the final results would not be confounded by discrepancies in data handling or model training procedures. By maintaining a unified structure across all models, the design enables a direct and interpretable comparison of their relative strengths and weaknesses.

All datasets were preprocessed through a standardized procedure. Missing values were replaced with zeros, which ensured that no samples were excluded from the training process due to incomplete feature information. This choice was made deliberately, as excluding samples would reduce the effective sample size and potentially bias the experimental results. Categorical features were transformed into a numerical representation. To maintain a consistent feature naming convention. After preprocessing, each dataset was randomly divided into training and test sets with a fixed split ratio of 7:3. By averaging results across repeated trials, we obtained performance measures that more faithfully reflect the expected generalization behavior of each model.

The evaluation of models was carried out with two widely recognized classification metrics. Because many of the datasets exhibit varying degrees of class imbalance, the F1-score was chosen as the primary metric of interest. In the current implementation, the default binary F1-score provided, which quantifies the harmonic mean of precision and recall for binary classification tasks. This metric is particularly informative in imbalanced settings, as it penalizes models that perform well on the majority class but fail to recognize the minority class. Accuracy was also reported as a secondary metric, measuring the proportion of correctly classified samples among the total. While accuracy provides a general sense of correctness, it can be misleading when the class distribution is highly skewed. For this reason, the subsequent analysis focuses more heavily on the F1-score, with accuracy presented mainly as a complementary indicator.

A broad spectrum of baseline classifiers was included to contextualize the performance of the proposed model. Among the traditional methods, we considered single decision trees trained with the Gini criterion, as well as Extremely Randomized Trees (Extrees) and Random Forests (RF), which introduce additional randomness to reduce variance and improve robustness. Gradient boosting methods were also included, specifically XGBoost (XGb) and LightGBM (LGBM), both of which are widely regarded as state-of-the-art ensemble methods for structured data. They provide useful reference points for situating the performance of NDLT in relation to classical decision tree families. All baselines were executed with default hyperparameters provided by their respective libraries, thereby ensuring that the comparison highlights structural differences among models rather than the effects of extensive parameter tuning. Each model was trained and evaluated on the same train/test splits, and the process was repeated ten times with distinct random seeds. The reported results correspond to the average F1-score and accuracy across these repetitions, reducing the impact of outlier runs.

For the proposed NDLT, the model was configured as a single decision tree with the maximum depth fixed at 10, a setting chosen to prevent overfitting and to keep the structure interpretable. The next-depth lookahead mechanism was constrained by the global minimum leaf size requirement on both child nodes, which avoids trivial splits and stabilizes the partitioning process. The feature sampling ratio was set to $r_t=1$, such that all available features were retained at each split without subsampling. The process of split selection was further controlled by two hyperparameters: $\beta$, which determines the number of top-ranked features considered for splitting, and $\gamma$, which specifies the number of candidate thresholds per feature. In this study, we evaluated three distinct combinations of $(\beta, \gamma)$, namely $(1,3)$, $(3,3)$, and $(3,5)$. These combinations were chosen to capture different trade-offs between the breadth of feature exploration and the granularity of threshold selection, thereby providing insight into how these factors influence the overall discriminative capacity of the model.

The parameter $w_{2}$ was treated as the central experimental factor. This parameter scales the relative importance of the immediate impurity reduction at the current node versus the potential improvement observable in the next depth level. To analyze its effect, we performed a systematic grid search over values ranging from $0.1$ to $1.0$ in increments of $0.1$. For each value of $w_{2}$, a new NDLT instance was trained and evaluated, and the procedure was repeated across ten random seeds. This design produced averaged performance measures for every configuration, enabling us to assess how the trade-off mechanism influences the balance between greedy and lookahead splitting strategies.

In summary, the experimental design ensures comparability across models by unifying preprocessing steps, evaluation metrics, and repeated train/test splitting. The inclusion of a wide array of baselines provides context for situating the performance of NDLT, while the exploration of different $(\beta, \gamma)$ settings and the systematic variation of $w_{2}$ allow us to probe the internal dynamics of the proposed algorithm. By averaging results across multiple runs, the reported outcomes reflect stable performance characteristics rather than artifacts of specific random splits. The NDLT itself was realized through a custom decision tree class that integrates the best-split routine and the trade-off weighting scheme. The outputs of each experiment consist of F1-score and accuracy values, and the subsequent section reports the averaged results obtained across all experimental settings.

\subsection{Performance Analysis of NDLT}

\begin{table}[htbp]
\caption{Comparison of F1 results of the NDLT and traditional models on various datasets with $\delta = 0.99$} 
\label{tab:NDLT_results_F1}
\centering
\resizebox{\linewidth}{!}{
\begin{tabular}{ccccccccccccccccc}
    \hline
    \textbf{NDLT} $\boldsymbol{(\beta,\:\gamma)}$ & \multicolumn{3}{c}{\textbf{NDLT (1, 3)}} & \multicolumn{3}{c}{\textbf{NDLT (3, 3)}} & \multicolumn{3}{c}{\textbf{NDLT  (3, 5)}} & \multicolumn{2}{c}{\textbf{Random Forest}} & \multicolumn{5}{c}{\textbf{Default Parameter}} \\ \cmidrule(lr){1-1} \cmidrule(lr){2-4}  \cmidrule(lr){5-7} \cmidrule(lr){8-10} \cmidrule(lr){11-12} \cmidrule(lr){13-17}
    \textbf{Dataset} & \textbf{MEAN} & \makecell{\textbf{Average} \\ \textbf{0.5-0.9}} & \textbf{max} & \textbf{MEAN} & \makecell{\textbf{Average} \\ \textbf{0.5-0.9}} & \textbf{max} & \textbf{MEAN} & \makecell{\textbf{Average} \\ \textbf{0.5-0.9}} & \textbf{max} & \makecell{\textbf{100 fixed} \\ \textbf{depth}} & \makecell{\textbf{Infinity} \\ \textbf{depth}} & \textbf{DT} &  \textbf{Extrees} & \textbf{RF} & \textbf{XGb} & \textbf{LGBM} \\
    
    \midrule   
    \textbf{Bank}       & 0.471  &  0.473  &  0.484  &  0.473  &  0.482  &  0.489  &  0.409  &  0.456  &  0.483  &  0.485  &  0.519  &  0.449  &  0.443  &  0.485  &  \textbf{0.536}  &  \textbf{0.536}   \\
    \textbf{Blood}      & 0.422  &  0.428  &  0.434  &  0.405  &  0.422  &  \textbf{0.463}  &  0.305  &  0.342  &  0.455  &  0.402  &  0.403  &  0.365  &  0.366  &  0.380  &  0.392  &  0.383   \\
    \textbf{Breast}     & 0.909  &  0.909  &  0.912  &  0.907  &  0.910  &  0.911  &  0.908  &  0.911  &  0.912  &  0.947  &  0.948  &  0.898  &  0.951  &  0.946  &  0.948  &  \textbf{0.955}   \\
    \textbf{Kidney}     & 0.919  &  0.921  &  0.927  &  0.917  &  0.918  &  0.927  &  0.921  &  0.922  &  0.927  &  0.944  &  0.971  &  0.918  &  \textbf{0.985}  &  0.977  &  0.973  &  0.977   \\
    \textbf{Voting}     & 0.918  &  0.916  &  0.926  &  0.918  &  0.916  &  0.926  &  0.924  &  0.920  &  \textbf{0.933}  &  0.956  &  0.955  &  0.923  &  0.946  &  0.949  &  0.941  &  0.942   \\
    \textbf{Heart}      & 0.643  &  0.663  &  0.687  &  0.650  &  0.661  &  0.682  &  0.650  &  0.665  &  0.695  &  \textbf{0.769}  &  0.759  &  0.669  &  0.666  &  0.753  &  0.727  &  0.715   \\
    \textbf{Horse}      & 0.766  &  0.770  &  0.780  &  0.765  &  0.767  &  0.785  &  0.760  &  0.766  &  0.785  &  \textbf{0.836}  &  0.825  &  0.785  &  0.802  &  0.818  &  0.808  &  0.806   \\
    \textbf{Iranian}    & 0.905  &  0.904  &  0.910  &  0.906  &  0.906  &  0.911  &  0.905  &  0.905  &  0.910  &  0.850  &  0.853  &  0.907  &  0.827  &  0.913  &  \textbf{0.926}  &  0.913   \\
    \textbf{Ozone}      & 0.453  &  0.456  &  0.465  &  0.447  &  0.454  &  0.466  &  0.449  &  0.454  &  0.464  &  \textbf{0.494}  &  0.490  &  0.472  &  0.473  &  0.491  &  0.506  &  0.489   \\
    \textbf{Raisin}     & 0.820  &  0.822  &  0.824  &  0.821  &  0.823  &  0.829  &  0.827  &  0.828  &  0.835  &  \textbf{0.862}  &  0.857  &  0.808  &  0.853  &  0.857  &  0.850  &  0.853   \\
    \textbf{Regensburg} & 0.907  &  0.906  &  0.910  &  0.908  &  0.908  &  0.913  &  0.907  &  0.908  &  0.913  &  \textbf{0.928}  &  0.926  &  0.904  &  0.830  &  0.913  &  0.926  &  0.913   \\
    \textbf{Rice}       & 0.915  &  0.915  &  0.917  &  0.917  &  0.919  &  0.921  &  0.919  &  0.921  &  0.923  &  \textbf{0.934}  &  0.931  &  0.902  &  0.929  &  0.931  &  0.927  &  0.928   \\
    \textbf{Spambase}   & 0.896  &  0.896  &  0.899  &  0.895  &  0.896  &  0.898  &  0.898  &  0.901  &  0.905  &  0.939  &  0.940  &  0.883  &  0.943  &  0.938  &  0.941  &  \textbf{0.944}   \\
    \midrule
    \textbf{Average}    & 0.765  &  0.768  &  0.775  &  0.764  &  0.768  &  0.779  &  0.752  &  0.761  &  0.780  &  0.796  &  0.798  &  0.752  &  0.772  &  0.791  &  \textbf{0.795}  &  0.794   \\
    \bottomrule
\end{tabular}
}
\end{table}

\begin{table}[htbp]
\caption{Comparison of Accuracy results of the NDLT and traditional models on various datasets with $\delta = 0.99$}
\label{tab:NDLT_results_Acc}
\centering
\resizebox{\linewidth}{!}{
\begin{tabular}{ccccccccccccccccc}
\hline
\textbf{NDLT} $\boldsymbol{(\beta,\:\gamma)}$ & \multicolumn{3}{c}{\textbf{NDLT (1, 3)}} & \multicolumn{3}{c}{\textbf{NDLT (3, 3)}} & \multicolumn{3}{c}{\textbf{NDLT  (3, 5)}} & \multicolumn{2}{c}{\textbf{Random Forest}} & \multicolumn{5}{c}{\textbf{Default Parameter}} \\ 
\cmidrule(lr){1-1} \cmidrule(lr){2-4}  \cmidrule(lr){5-7} \cmidrule(lr){8-10} \cmidrule(lr){11-12} \cmidrule(lr){13-17}
\textbf{Dataset} & \textbf{MEAN} & \makecell{\textbf{Average} \\ \textbf{0.5-0.9}} & \textbf{max} & \textbf{MEAN} & \makecell{\textbf{Average} \\ \textbf{0.5-0.9}} & \textbf{max} & \textbf{MEAN} & \makecell{\textbf{Average} \\ \textbf{0.5-0.9}} & \textbf{max} & \makecell{\textbf{100 fixed} \\ \textbf{depth}} & \makecell{\textbf{Infinity} \\ \textbf{depth}} & \textbf{DT} &  \textbf{Extrees} & \textbf{RF} & \textbf{XGb} & \textbf{LGBM} \\

\midrule
\textbf{Bank}       & 0.898  &  0.898  &  0.899  &  0.898  &  0.899  &  0.899  &  0.897  &  0.899  &  0.900  &  0.905  &  0.906  &  0.869  &  0.900  &  0.905  &  0.906  &  \textbf{0.907}   \\
\textbf{Blood}      & 0.732  &  0.734  &  0.736  &  0.739  &  0.741  &  \textbf{0.748}  &  0.731  &  0.737  &  0.755  &  0.755  &  0.756  &  0.712  &  0.730  &  0.736  &  0.745  &  0.747   \\
\textbf{Breast}     & 0.933  &  0.933  &  0.935  &  0.932  &  0.934  &  0.935  &  0.932  &  0.935  &  0.936  &  0.958  &  0.963  &  0.929  &  0.963  &  0.961  &  0.963  &  \textbf{0.968}   \\
\textbf{Kidney}     & 0.940  &  0.942  &  0.946  &  0.938  &  0.939  &  0.946  &  0.941  &  0.942  &  0.946  &  0.944  &  0.979  &  0.941  &  \textbf{0.991}  &  0.983  &  0.980  &  0.980   \\
\textbf{Voting}     & 0.939  &  0.937  &  0.944  &  0.939  &  0.937  &  0.944  &  0.943  &  0.940  &  0.950  &  \textbf{0.967}  &  \textbf{0.967}  &  0.947  &  0.960  &  0.963  &  0.957  &  0.958   \\
\textbf{Heart}      & 0.781  &  0.787  &  0.800  &  0.784  &  0.788  &  0.801  &  0.787  &  0.794  &  0.816  &  \textbf{0.866}  &  0.860  &  0.803  &  0.826  &  0.860  &  0.840  &  0.836   \\
\textbf{Horse}      & 0.824  &  0.829  &  0.838  &  0.822  &  0.826  &  0.840  &  0.822  &  0.824  &  0.839  &  0.836  &  \textbf{0.878}  &  0.785  &  0.802  &  0.818  &  0.808  &  0.806   \\
\textbf{Iranian}    & 0.924  &  0.923  &  0.927  &  0.925  &  0.924  &  0.928  &  0.924  &  0.924  &  0.928  &  0.954  &  \textbf{0.956}  &  0.924  &  0.865  &  0.932  &  0.941  &  0.931   \\
\textbf{Ozone}      & 0.949  &  0.950  &  0.951  &  0.948  &  0.949  &  0.951  &  0.947  &  0.948  &  0.951  &  0.960  &  0.960  &  0.954  &  0.960  &  0.960  &  \textbf{0.961}  &  0.959   \\
\textbf{Raisin}     & 0.817  &  0.820  &  0.822  &  0.819  &  0.821  &  0.827  &  0.824  &  0.824  &  0.830  &  \textbf{0.859}  &  0.855  &  0.812  &  0.850  &  0.855  &  0.850  &  0.853   \\
\textbf{Regensburg} & 0.924  &  0.924  &  0.927  &  0.926  &  0.926  &  0.930  &  0.924  &  0.926  &  0.930  &  0.940  &  \textbf{0.941}  &  0.926  &  0.868  &  0.930  &  \textbf{0.941}  &  0.931   \\
\textbf{Rice}       & 0.902  &  0.902  &  0.905  &  0.904  &  0.907  &  0.909  &  0.907  &  0.910  &  0.912  &  \textbf{0.924}  &  0.921  &  0.887  &  0.918  &  0.921  &  0.916  &  0.918   \\
\textbf{Spambase}   & 0.918  &  0.918  &  0.921  &  0.918  &  0.918  &  0.920  &  0.920  &  0.922  &  0.925  &  0.950  &  0.953  &  0.908  &  0.954  &  0.952  &  0.954  &  \textbf{0.956}   \\
\midrule
\textbf{Average}    & 0.883  &  0.884  &  0.888  &  0.884  &  0.885  &  0.891  &  0.884  &  0.886  &  0.894  &  0.877  &  \textbf{0.915}  &  0.880  &  0.856  &  0.874  &  0.876  &  0.875   \\
\bottomrule
\end{tabular}
}
\end{table}

In Tables~\ref{tab:NDLT_results_F1} and \ref{tab:NDLT_results_Acc}, we report the F1-scores and accuracies obtained by applying the proposed NDLT model and several conventional tree-based models to a diverse collection of datasets. For the NDLT, three evaluation settings were considered for each configuration of $(\beta,\gamma)$: \emph{Mean}, \emph{Average 0.5--0.9}, and \emph{Max}. These settings were designed to provide complementary perspectives on the stability and potential of the proposed model, thereby facilitating a fair comparison with traditional decision tree families such as single CART, Extremely Randomized Trees, Random Forests, and gradient boosting ensembles.

The \emph{Mean} result corresponds to the average performance obtained when training the NDLT ten times for each $w_{2} \in \{0.1,0.2,\dots,1.0\}$. The \emph{Average 0.5--0.9} result is defined analogously, but restricted to the subset $w_{2} \in \{0.5,0.6,\dots,0.9\}$. Finally, the \emph{Max} result denotes the best performance achieved among all $w_{2}$ values in $\{0.1,0.2,\dots,1.0\}$, again based on ten repeated runs per setting. Taken together, these three summaries highlight not only the typical behavior of NDLT across a full range of trade-off weights but also its robustness in a mid-range regime and its best-case potential relative to established baselines.

In terms of overall performance, the NDLT consistently outperformed the conventional single DT across most datasets, with particularly strong advantages observed in the F1 score. When compared to ensemble models such as XGBoost and LightGBM, the accuracy of NDLT was slightly lower, but the difference was marginal. In contrast, the F1 score was generally comparable to or even higher than that of the ensembles, indicating that the NDLT is able to secure strong discriminative capacity despite its single-tree structure.

The analysis further reveals that the behavior of the models differs depending on the evaluation metric. In datasets characterized by severe class imbalance, such as Bank, Blood and Ozone, the discrepancy between Accuracy and F1 score was substantial, showing that accuracy alone is insufficient to properly assess model performance. This confirms that adopting the F1 score as the primary evaluation metric in this study was an appropriate choice, as it more effectively reflects the model’s ability to handle imbalanced data.

With respect to the parameter $w_{2}$, stable performance was observed across all three evaluation settings: \emph{Mean}, \emph{Average 0.5--0.9}, and \emph{Max}. In particular, the \emph{Average 0.5--0.9} setting demonstrates that NDLT maintains consistent performance across a mid-range of $w_{2}$ values rather than depending on a single configuration, thereby serving as an indicator of robustness. Moreover, under the \emph{Max} criterion, NDLT achieved results that were on par with, or superior to, those of tuned ensemble models such as XGBoost and LightGBM on several datasets.

The results also highlight performance variations depending on dataset characteristics. For high-dimensional datasets such as Kidney, Voting, and Spambase, the NDLT performed comparably to or better than ensemble models, suggesting that the proposed method is capable of effectively capturing complex patterns within a single-tree framework. In contrast, for relatively simple datasets such as Breast and Rice, LightGBM and XGBoost achieved slightly higher performance, although the gap remained limited. Traditional tree-based models such as DT and Extrees were consistently outperformed by NDLT, supporting the effectiveness of its structural optimization and lookahead-based split selection mechanism.

Taken together, these findings indicate that the NDLT achieves comparable or superior performance relative to ensemble methods in terms of F1 score, while establishing a clear advantage over single-tree baselines. Although the accuracy of NDLT was somewhat lower than that of ensembles, the difference was within approximately 1\% on average, and in imbalanced datasets the F1 score provided a more reliable evaluation of model performance than accuracy alone.

\begin{figure}[ht!]
    \centering
    \includegraphics[width=\linewidth]{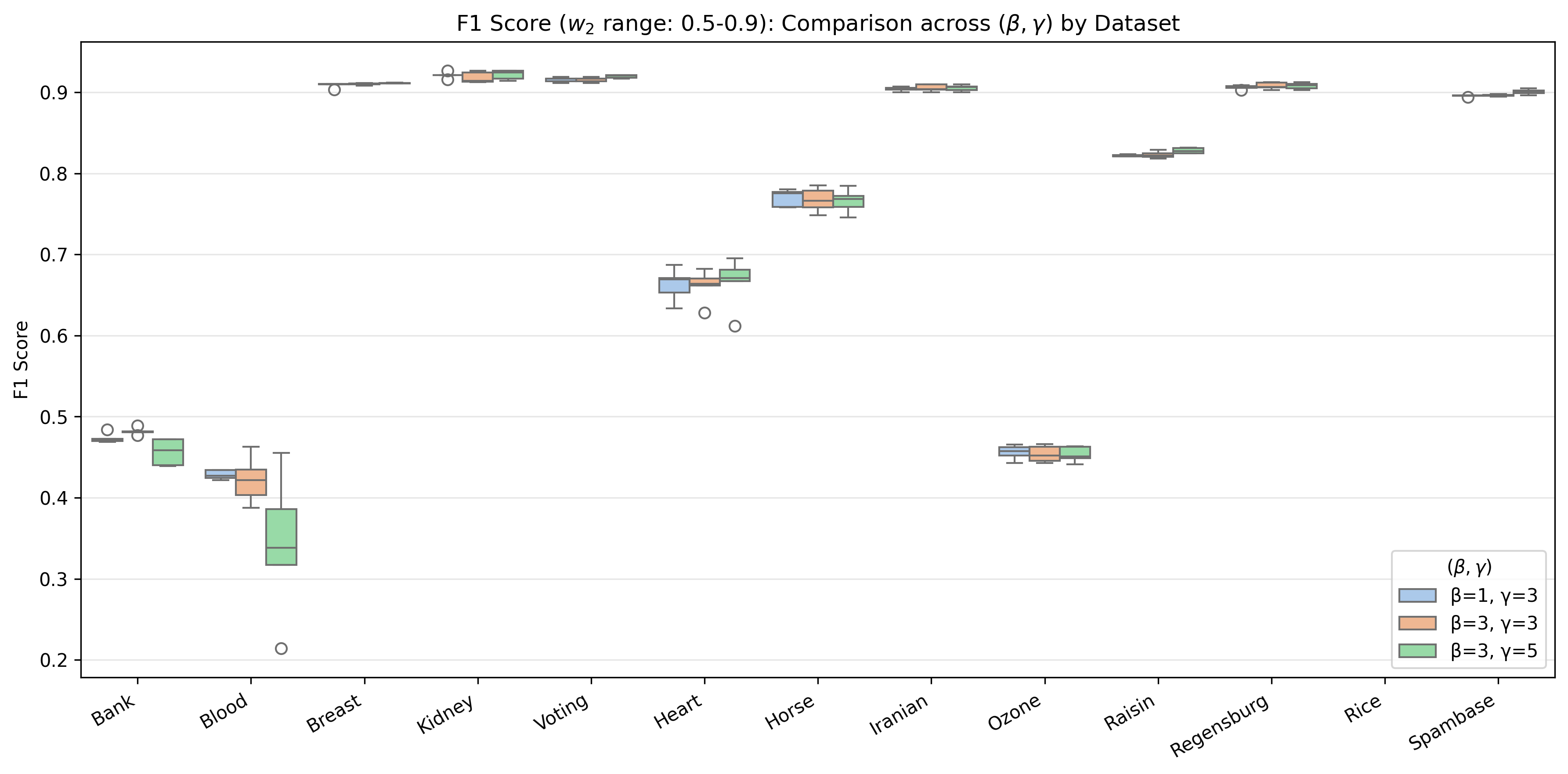}
    \caption{F1-score comparison of the proposed NDLT across UCI datasets under three $(\beta,\gamma)$ settings $\{(1,3),(3,3),(3,5)\}$ with $w_{2}\in[0.5,0.9]$. Each box shows the distribution of F1 scores over repeated runs for a given dataset. Increasing $\beta$ (top-feature pool) or $\gamma$ (candidate thresholds per feature) generally yields equal or slightly higher medians—with clearer gains on Heart, Horse, and Ozone—while improvements are limited on Bank/Blood where imbalance and coarse features restrict useful split points.}
  \label{fig:ndlt_f1_box_by_dataset}
\end{figure}

Figure~\ref{fig:ndlt_f1_box_by_dataset} summarizes the dataset-wise distributions of F1 for the proposed NDLT under three $(\beta,\gamma)$ settings, $\{(1,3),(3,3),(3,5)\}$, while constraining the weighting parameter to $w_{2} \in[0.5,0.9]$. In this configuration, $\beta$ denotes the size of the top-feature shortlist admitted to the inner best-split routine and $\gamma$ is the number of candidate thresholds per feature examined by the next-depth lookahead. Each box reflects repeated runs with different random seeds for a fixed dataset and $(\beta,\gamma)$, thus exposing not only central tendency but also the seed-level variability induced by feature subsampling and threshold proposals. 

A clear stratification emerges across datasets. Kidney, Voting, Iranian, Regensburg, Rice, and Spambase concentrate in the high-F1 band (upper 0.8 to 0.9+), with compact interquartile ranges and few outliers; this pattern indicates that NDLT identifies stable, near-saturated partitions with little sensitivity to seed-level perturbations. Raisin forms a mid–high plateau with consistently narrow boxes, suggesting well-separated clusters with limited ambiguity. Heart and Horse occupy a mid-range regime where medians are lower and spreads are wider than the aforementioned easy group; here the quality of local partitions depends more visibly on how many strong features and thresholds are made available to the lookahead. By contrast, Bank, Blood, Breast, and Ozone exhibit clearly lower medians, delineating the challenging end of the spectrum in this study.

Fixing $\gamma$ and increasing $\beta$ from $1$ to $3$ (comparing $(1,3)$ to $(3,3)$) generally elevates medians and mildly contracts the boxes for Heart, Horse, and Ozone, while leaving the already-easy group nearly unchanged. This behavior is consistent with the mechanism of the algorithm: enlarging the shortlist of strong features reduces the chance that a myopic parent-level choice steers the inner search toward locally inferior splits, thereby allowing the next-depth evaluation to discover better-balanced partitions. On Bank and Blood, however, the effect of enlarging $\beta$ is limited. Bank Marketing is severely imbalanced (positive rate $\approx 11.7\%$) and heavily categorical, which restricts the effective granularity of informative cuts even when multiple strong features are considered; Blood Transfusion is small (748 observations) with only four predictors, so expanding the shortlist yields little diversity in admissible splits and leaves the model comparatively sensitive to seed variation.

Holding $\beta$ fixed and increasing $\gamma$ from $3$ to $5$ (contrasting $(3,3)$ with $(3,5)$) produces monotone or at least non-degrading trends in nearly all datasets, with the clearest benefits again on Heart, Horse, and Ozone and a modest improvement on Raisin. The interpretation is that denser threshold settings expose finer candidate midpoints/quantiles that, subject to the $min\_samples\_leaf$ constraint, yield more balanced children and slightly higher purity reduction at the sidewise evaluation stage. Gains are not dramatic—\\diminishing returns arise once the major split locations have been sampled—but they are systematic and accompanied by reduced dispersion, which is desirable from a robustness standpoint.

The dataset-specific geometry helps explain the difficult cases. Bank’s combination of class imbalance and many categorical fields depresses F1 relative to accuracy-oriented summaries because positive recall is intrinsically costly without sufficient granular cut points. Blood’s small $n$ and very low dimensionality constrain the search space and amplify stochastic fluctuations across seeds. Ozone exhibits noisy, overlapping class boundaries; both $\beta$ and $\gamma$ help a little by refining local partitions, yet the intrinsic diffuseness of the boundary sets a ceiling on achievable F1. Breast shows a lower median with a conspicuously wide spread, including an outlier run with markedly poor F1; under the settings $w_{2} \in[0.5,0.9]$ and $\gamma\in\{3,5\}$, the threshold set may still be too sparse to consistently carve high-contrast splits among correlated radiomic predictors, suggesting that either a larger $\gamma$ or a slightly different weighting window could stabilize outcomes.

Taken together, the figure indicates three practical lessons. First, modest increases in either $\beta$ or $\gamma$ seldom harm F1 and tend to help exactly where the boundary is ambiguous; improvements are incremental but reliable and often accompanied by lower variability. Second, the reasons for weak performance are structural and dataset-dependent—imbalance and categorical sparsity (Bank), sample/feature scarcity (Blood), noisy overlap (Ozone), and under-segmentation under current settings (Breast)—so budget tuning should be targeted rather than global. Third, because computational cost scales roughly linearly with the number of shortlisted features and tested thresholds per node, $(\beta,\gamma)=(3,5)$ offers a favorable accuracy–cost compromise for mixed tabular regimes, whereas $(1,3)$ is adequate for the easy group (Kidney, Voting, Iranian, Regensburg, Rice, Spambase). Within the scope of Fig.~\ref{fig:ndlt_f1_box_by_dataset}, these observations support the claim that NDLT maintains stable, high F1 on diverse data geometries and that judicious increases of $(\beta,\gamma)$ provide robust, if incremental, gains on more challenging datasets.

\section{Conclusion}
\label{sec:Conclusion}
This paper introduced the \emph{Next-Depth Lookahead Tree} (NDLT), a single-tree learning algorithm that augments the classical top-down decision-tree induction with a depth-$1$ lookahead. Instead of committing to a split solely by its immediate impurity reduction, NDLT evaluates each parent candidate by jointly considering its upper-node impurity and a lower-node proxy, obtained from a side-wise, one-step localized search with explicit admissibility checks. This design preserves the interpretability of white-box trees while mitigating well-known limitations of purely greedy induction such as myopic feature choices and susceptibility to locally suboptimal early splits.

Across thirteen heterogeneous UCI datasets, the empirical study demonstrates that NDLT achieves \emph{stable and competitive} performance. In Fig.~\ref{fig:ndlt_f1_box_by_dataset}, the distribution of F1 scores shows a clear stratification: on \emph{Kidney, Voting, Iranian, Regensburg, Rice}, and \emph{Spambase}, NDLT attains high medians with narrow IQRs across all tested $(\beta,\gamma)$ settings, indicating near-saturated partitions and robustness to seed-level perturbations. On tasks with greater ambiguity---notably \emph{Heart}, \emph{Horse}, and \emph{Ozone}---increasing either the top-feature shortlist size $\beta$ or the threshold budget $\gamma$ yields modest but consistent improvements, often accompanied by reduced dispersion. The difficult cases (\emph{Bank}, \emph{Blood}, \emph{Breast}, \emph{Ozone}) align with structural factors such as severe imbalance, low dimensionality, categorical sparsity, and noisy or overlapping boundaries. These observations are consistent with the averaged results in Tables~\ref{tab:NDLT_results_F1} and \ref{tab:NDLT_results_Acc}: NDLT clearly outperforms a conventional single-tree baseline and is frequently competitive with ensemble methods in terms of F1, while trailing those ensembles by only a small margin in accuracy.

The ablation on the lookahead settings suggests practical guidance. When $\gamma$ is fixed, enlarging $\beta$ from $1$ to $3$ reliably helps on the ambiguous group while leaving the easy group nearly unchanged; conversely, at fixed $\beta$, raising $\gamma$ from $3$ to $5$ produces monotone or non-degrading trends with visible variance reduction on several datasets. Since the computational cost scales roughly linearly with the number of shortlisted features and candidate thresholds per node, $(\beta,\gamma)=(3,5)$ presents a favorable accuracy--cost compromise for mixed tabular regimes, whereas $(1,3)$ is sufficient for the easy group identified above. Moreover, the weighting factor $w_{2}$ exhibits stable behavior over the mid-range $[0.5,0.9]$, indicating that NDLT does not hinge on a narrow hyperparameter setting to realize its gains.

NDLT remains a \emph{single} tree and thus retains path-wise transparency, rule-level auditability, and low deployment overhead, features that are often traded away by large ensembles. At the same time, the depth-$1$ lookahead reduces the risk that early greedy commitments lock the tree into poor partitions. Taken together, these properties make NDLT an attractive option when interpretability and engineering simplicity are first-class requirements but one still seeks performance approaching that of modern ensembles.

There are several avenues for future work. On the algorithmic side, one could (i) learn $(\beta,\gamma)$ and the weighting $w_{2}$ adaptively per node via validation-guided or Bayesian schemes; (ii) extend the lookahead beyond a single depth with budget-aware pruning; (iii) incorporate oblique (linear) or categorical-aware splits while keeping the same lower-node evaluation principle; and (iv) couple the lookahead with principled post-hoc pruning and probability calibration. On the theoretical side, deriving generalization and consistency guarantees for the combined upper/lower objective would sharpen our understanding of when and why lookahead helps. Finally, systems work---including GPU-accelerated threshold evaluation and streaming or online variants---could further reduce latency and extend NDLT to real-time decisioning. We hope these directions encourage broader adoption of lookahead strategies in interpretable tree learning.

\newpage
\bibliographystyle{elsarticle-num}  

\begin{thebibliography}{10}
\expandafter\ifx\csname url\endcsname\relax
  \def\url#1{\texttt{#1}}\fi
\expandafter\ifx\csname urlprefix\endcsname\relax\def\urlprefix{URL }\fi
\expandafter\ifx\csname href\endcsname\relax
  \def\href#1#2{#2} \def\path#1{#1}\fi

\bibitem{izza2020explain}
Y.~Izza, A.~Ignatiev, J.~Marques-Silva, On explaining decision trees, arXiv preprint arXiv:2010.11034 (2020).

\bibitem{costa2023survey}
V.~G. Costa, C.~E. Pedreira, Recent advances in decision trees: An updated survey, Artificial Intelligence Review 56~(5) (2023) 4765--4800.

\bibitem{kazemitabar2017variable}
J.~Kazemitabar, A.~Amini, A.~Bloniarz, A.~Talwalkar, Variable importance using decision trees, in: Advances in Neural Information Processing Systems, Vol.~30, 2017.

\bibitem{ahmed2023turnover}
S.~R. Ahmed, A.~K. Ahmed, S.~J. Jwmaa, Analyzing the employee turnover by using decision tree algorithm, in: 2023 5th International Congress on Human-Computer Interaction, Optimization and Robotic Applications (HORA), IEEE, 2023, pp. 1--4.

\bibitem{wang2024fruit}
Z.~Wang, Fruit and vegetable image recognition based on multiple tree models: Applications of random forest, xgboost and decision tree, Technology and Engineering 1~(9) (2024) 1--12.

\bibitem{amruth2024cloud}
A.~Amruth, R.~Ramanan, R.~Paul, C.~Vimal, B.~Beena, Cloud based big data solution for cancer classification: Using databricks on large scale genomic data, in: 2024 1st International Conference on Communications and Computer Science (InCCCS), IEEE, 2024, pp. 1--6.

\bibitem{choi2017manufacturing}
S.~Choi, L.~Battulga, A.~Nasridinov, K.-H. Yoo, A decision tree approach for identifying defective products in the manufacturing process, International Journal of Contents 13~(2) (2017) 57--65.

\bibitem{strobl2009recursive}
C.~Strobl, J.~Malley, G.~Tutz, An introduction to recursive partitioning: rationale, application, and characteristics of classification and regression trees, bagging, and random forests, Psychological Methods 14~(4) (2009) 323.

\bibitem{kozak2019dt}
J.~Kozak, Decision tree and ensemble learning based on ant colony optimization, in: Proceedings of the International Conference on Artificial Intelligence, 2019, pp. 1--6.

\bibitem{olaru2003fuzzy}
C.~Olaru, L.~Wehenkel, A complete fuzzy decision tree technique, Fuzzy Sets and Systems 138~(2) (2003) 221--254.

\bibitem{rouhi2020feature}
A.~Rouhi, H.~Nezamabadi-Pour, Feature selection in high-dimensional data, in: Optimization, Learning, and Control for Interdependent Complex Networks, Springer, 2020, pp. 85--128.

\end{thebibliography}


\end{document}